\documentclass[lettersize,journal]{IEEEtran}

\usepackage{amsmath}   
\usepackage{amssymb}   
\usepackage{bm}        

\usepackage{amsfonts}
\usepackage{algorithm} 
\usepackage{algorithmic}
\usepackage{array}
\usepackage[caption=false,font=normalsize,labelfont=sf,textfont=sf]{subfig}
\usepackage{textcomp}
\usepackage{stfloats}
\usepackage{url}
\usepackage{verbatim}
\usepackage{graphicx}
\usepackage{array}
\usepackage[caption=false,font=normalsize,
   labelfont=sf,textfont=sf]{subfig}
\usepackage{textcomp}
\usepackage{stfloats}
\usepackage{url}
\usepackage{verbatim}
\usepackage{graphicx}
\usepackage{cite}
\usepackage{generic}
\usepackage{makecell}
\usepackage{epstopdf}
\usepackage[table,xcdraw]{xcolor}
\usepackage[normalem]{ulem}
\usepackage{booktabs}
\usepackage{enumitem}
\usepackage{multirow}
\usepackage{afterpage}
\useunder{\uline}{\ul}{}
\usepackage{tabularx} 
\usepackage{ragged2e} 
\usepackage{amsmath}

\def\BibTeX{{\rm B\kern-.05em{\sc i\kern-.025em b}\kern-.08em
    T\kern-.1667em\lower.7ex\hbox{E}\kern-.125emX}}
\usepackage{balance}
\usepackage{hyperref}

\begin{document}
\title{Heterogeneous Multi-Expert Reinforcement Learning for Long-Horizon Multi-Goal Tasks in Autonomous Forklifts}

\author{Yun Chen, Bowei Huang, Fan Guo, Kang Song

\thanks{This work was supported in part by the National Key Research and Development Program of China under Grant 2022YFE0100100, the Tianjin Municipal Science and Technology Program under Grant 19ZXZNGX00050, the State Key Laborator of Engines at Tianjin University under Grant K2022-13 and the Research and Innovation Project for Postgraduates in Tianjin under Grant 2022BKYZ041. (Corresponding author: Kang Song.)

Yun Chen, Kang Song, Fan Guo and Bowei Huang are with the State Key Laboratory of Engines, Tianjin University, Tianjin 300072, China. (e-mail: cyun@tju.edu.cn; songkangtju@tju.edu.cn).
}}
\markboth{}%
{How to Use the IEEEtran \LaTeX \ Templates}

\maketitle

\begin{abstract}
Autonomous mobile manipulation in unstructured warehouses requires a balance between efficient large-scale navigation and high-precision object interaction. Traditional end-to-end learning approaches often struggle to handle the conflicting demands of these distinct phases. Navigation relies on robust decision-making over large spaces, while manipulation needs high sensitivity to fine local details. Forcing a single network to learn these different objectives simultaneously often causes optimization interference, where improving one task degrades the other. To address these limitations, we propose a Heterogeneous Multi-Expert Reinforcement Learning (HMER) framework tailored for autonomous forklifts. HMER decomposes long-horizon tasks into specialized sub-policies controlled by a Semantic Task Planner. This structure separates macro-level navigation from micro-level manipulation, allowing each expert to focus on its specific action space without interference. The planner coordinates the sequential execution of these experts, bridging the gap between task planning and continuous control. Furthermore, to solve the problem of sparse exploration, we introduce a Hybrid Imitation-Reinforcement Training Strategy. This method uses expert demonstrations to initialize the policy and Reinforcement Learning for fine-tuning. Experiments in Gazebo simulations show that HMER significantly outperforms sequential and end-to-end baselines. Our method achieves a task success rate of 94.2\% (compared to 62.5\% for baselines), reduces operation time by 21.4\%, and maintains placement error within 1.5 cm, validating its efficacy for precise material handling.
\end{abstract}

\begin{IEEEkeywords}
Keywords: Autonomous Forklift; Hierarchical Reinforcement Learning; Mobile Manipulation; Hybrid Training; Modality Decoupling.
\end{IEEEkeywords}

\section{Introduction}
In the rapidly evolving landscape of smart logistics, the demand for autonomous mobile manipulation—particularly involving heavy-duty machinery like forklifts—is increasing rapidly. Unlike Automatic Guided Vehicles (AGVs) that adhere to rigid, pre-defined paths, autonomous forklifts in modern warehouses must operate in unstructured environments and execute complex long-horizon tasks. These missions typically involve a continuous cycle: navigating through dynamic obstacles, visually searching for cargo, picking it up, and stacking it with centimeter-level precision. This dual requirement poses a significant robotic control challenge: the system must simultaneously master large-scale spatial navigation and fine-grained object interaction, integrating sequential decision-making with complex vehicle dynamics.
However, developing a robust control policy for such tasks remains an open problem due to three fundamental limitations in existing paradigms. First, traditional End-to-End Reinforcement Learning (RL) methods often succumb to optimization interference. Compelling a single network to simultaneously encode geometric navigation features and semantic manipulation cues results in gradient conflict, where the optimization of large-scale mobility dominates the learning of fine-grained interaction. Second, while sequential modular approaches mitigate this complexity by executing sub-tasks in series, they suffer from error propagation and irreversibility. Since these frameworks typically execute navigation and manipulation phases in isolation, minor state deviations accumulated during transit inevitably cascade into the manipulation phase. Lacking a high-level mechanism to dynamically revert states or re-plan, these systems cannot recover from such propagated errors, leading to task failure in unstructured environments. Third, although Imitation Learning (IL) offers a stable policy initialization, it is fundamentally restricted by the performance ceiling of the demonstrator. Pure supervised learning is confined to the distribution of the training data and precludes the discovery of superior control strategies through environmental interaction, thereby failing to achieve the sub-centimeter precision mandated by industrial operations.
To address these limitations, we propose a Heterogeneous Multi-Expert Reinforcement Learning (HMER) framework tailored for autonomous forklifts. Our core insight is to structurally decouple the complex long-horizon task into a hierarchy of specialized experts, thereby resolving the optimization interference problem. Specifically, we design a Navigation Expert that focuses on macro-level geometric robustness for path planning, and Manipulation Experts that focus on micro-level semantic details for precise interaction. These specialized skills are orchestrated by a high-level Semantic Task Planner, which acts as a closed-loop decision maker to bridge discrete task logic with continuous motor control.
Furthermore, to solve the trade-off between exploration efficiency and control precision, we introduce a unified Hybrid Imitation-Reinforcement Training Strategy. Training long-horizon tasks from scratch is computationally inefficient due to sparse rewards. We first utilize expert demonstrations to warm-start all expert policies via Behavioral Cloning (BC), establishing a baseline competency. Subsequently, we employ Reinforcement Learning (RL) to fine-tune these policies end-to-end. This refinement phase is critical: it enables the manipulation agents to learn subtle, non-intuitive control maneuvers that surpass the heuristic expert, achieving the high precision required for real-world deployment.
The main contributions of this paper are summarized as follows:
1.	A Heterogeneous Multi-Expert Reinforcement Learning Framework (HMER): Distinct from traditional end-to-end learning paradigms, we propose a hierarchical framework that fundamentally addresses the optimization interference problem in autonomous mobile manipulation. By explicitly decoupling macro-level navigation from micro-level manipulation, our approach allows each expert to specialize in its specific action space without conflicting gradients, ensuring robust performance.
2.	Hierarchical Semantic Task Planning: We design a high-level Semantic Task Planner to orchestrate the sequential execution of heterogeneous experts. Unlike rigid sequential methods, this state-aware mechanism enables the autonomous forklift to adaptively switch between navigation and stacking phases and autonomously recover from propagated errors via closed-loop decision making.
3.	Unified Hybrid Imitation-Reinforcement Training Strategy: To overcome the exploration sparsity of Reinforcement Learning (RL) and the precision bottlenecks of Imitation Learning (IL), we introduce a hybrid training paradigm. This strategy leverages expert demonstrations to warm-start policy networks and subsequently employs RL for residual refinement. This enables the system to achieve 1.5 cm placement precision, significantly surpassing the capabilities of heuristic baselines and meeting strict industrial safety standards.
The remainder of this paper is organized as follows: Section II reviews related work in autonomous mobile manipulation and hierarchical reinforcement learning. Section III formally formulates the long-horizon multi-goal task as a Markov Decision Process (MDP). Section IV details the proposed HMER framework, elaborating on the Semantic Task Planner and the decoupled expert policies, followed by the hybrid training strategy. Section V describes the high-fidelity simulation environment and implementation details. Section VI presents extensive experimental results, comparative analyses against baselines, and ablation studies. Finally, Section VII concludes the paper and outlines future directions.

\section{Related work}
Research on autonomous mobile manipulation has evolved significantly in recent years, broadly branching into three primary paradigms: classical planning, data-driven monolithic learning, and hierarchical approaches. In this section, we review the literature in these domains and position our work within the context of unstructured warehouse logistics.

\subsection{Task and Motion Planning (TAMP)}
Early attempts to solve mobile manipulation predominantly relied on classical Task and Motion Planning (TAMP). These approaches view navigation and manipulation as distinct geometric problems. TAMP methods \cite{kaelbling2011hierarchical, garrett2021integrated} integrate these by searching for a logically consistent sequence of symbolic actions coupled with continuous motion trajectories. To handle complex kinematic constraints, researchers have developed extensible interface layers \cite{srivastava2014combined} and optimization-based logic-geometric programming \cite{toussaint2015logic} to bridge symbolic planners with continuous solvers.

While theoretically sound in structured settings, TAMP methods suffer from a critical limitation: reliance on \textbf{``privileged information.''} They typically assume access to pre-built high-definition maps \cite{thrun2005probabilistic} or precise 3D object models \cite{simeonov2022neural} for collision checking. In the context of autonomous forklifts, this rigidity is problematic as warehouse layouts change dynamically with moving inventory. Unlike TAMP frameworks that require exact belief space estimation \cite{kaelbling2013integrated}, our approach relies purely on onboard sensors (LiDAR and RGB), allowing the agent to adapt to unstructured environments without perfect state estimation.

\subsection{End-to-End and Imitation Learning}
To overcome the dependency on manual modeling, researchers have turned to data-driven approaches. End-to-End Reinforcement Learning (RL) attempts to map raw sensor observations directly to control actions \cite{levine2016end}. While successful in stationary manipulation \cite{kalashnikov2018qt}, these methods struggle when applied to long-horizon mobile manipulation \cite{xia2020relmogen}. A primary challenge is \textbf{optimization interference}: optimizing a single monolithic network to simultaneously handle LiDAR-based navigation and Vision-based manipulation often leads to conflicting gradients \cite{yu2020gradient}, causing convergence failure or sub-optimal policies.

Alternatively, Imitation Learning (IL) bypasses the exploration difficulty by mimicking expert demonstrations. Algorithms ranging from classical Behavior Cloning to Generative Adversarial Imitation Learning (GAIL) \cite{ho2016generative} and the recent Diffusion Policy \cite{chi2023diffusion} have demonstrated impressive capabilities. However, IL suffers from two fundamental limitations in industrial settings: (1) \textbf{Covariate Shift}: slight deviations during execution lead to state distributions not covered in the training data, causing compounding errors \cite{ross2011reduction}; and (2) \textbf{Precision Bottleneck}: the policy is strictly upper-bounded by the quality of the dataset \cite{mandlekar2021matters} and lacks an online correction mechanism to achieve the sub-centimeter accuracy required for stacking. To address these, our work employs a hybrid strategy: we use IL for stable initialization and RL for fine-tuning, enabling the agent to exceed expert precision via residual learning.

\subsection{Hierarchical Reinforcement Learning (HRL)}
To address the temporal abstraction challenge, HRL decomposes complex decision-making into high-level planning and low-level skills \cite{barto2003recent}. Traditional approaches focus on sequencing pre-defined primitives, such as the Options framework \cite{sutton1999between} or FeUdal Networks \cite{vezhnevets2017feudal}. More recently, researchers have leveraged Large Language Models (LLMs) to perform high-level planning, as seen in SayCan \cite{ahn2022do} and VoxPoser \cite{huang2023voxposer}, which ground linguistic commands into robotic affordances.

However, a key gap remains in the \textbf{coordination of heterogeneous skills}. Most existing HRL frameworks treat sub-policies as homogeneous blocks, ignoring the distinct sensory requirements of different phases (e.g., Geometry vs. Semantics). Furthermore, learning these hierarchies from scratch is notoriously sample-inefficient. While hybrid methods like DAPG \cite{rajeswaran2018learning} or Residual Policy Learning \cite{johannink2019residual} combine demonstrations with RL, they are rarely applied to hierarchically decoupled systems for mobile manipulation. Our work fills this gap by proposing the HMER framework. By explicitly decoupling sensor modalities for Navigation and Manipulation experts and co-optimizing them via a hybrid pipeline, we effectively solve the ``cold start'' problem and achieve the centimeter-level precision required for autonomous forklifts.

\section{PROBLEM FORMULATION}
\begin{figure}[t]
    \centering
    \includegraphics[width=0.95\columnwidth]{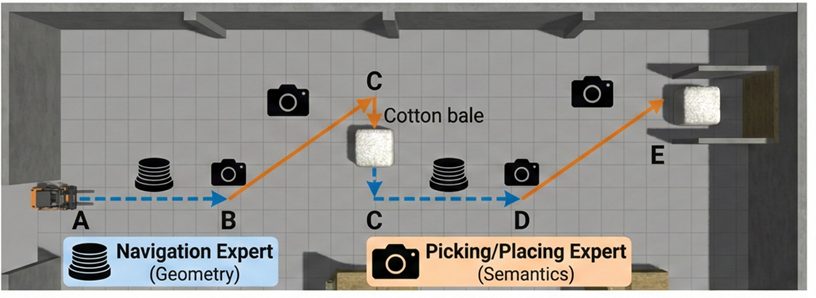} 
    
    \caption{\textbf{Illustration of the Long-Horizon Material Handling Task.} 
    The operational lifecycle is decomposed into four sequential phases: 
    (1) \textbf{Departure} from the docking station $\mathcal{P}_{start}$; 
    (2) \textbf{Search \& Pick} at the randomized cargo location $\mathcal{P}_{obj}$; 
    (3) \textbf{Transport} through dynamic obstacles; and 
    (4) \textbf{Precision Placement} at the target slot $\mathcal{P}_{goal}$. 
    The heterogeneous sensory inputs (LiDAR for navigation, RGB for manipulation) act as triggers for the state transitions.}
    
    \label{fig:task_overview}
\end{figure}
\subsection{Task Definition}
We formulate the autonomous mobile manipulation task as a long-horizon sequential decision-making problem within a stochastic warehouse environment. The operational lifecycle is decomposed into four distinct phases, as illustrated in Fig. \ref{fig:task_overview}: 
\begin{enumerate}
    \item \textbf{Departure:} The agent navigates from a docking station $\mathcal{P}_{start}$ to a designated transfer zone.
    \item \textbf{Search \& Pick:} The agent requires visual identification and precise alignment with a target cargo initialized at a stochastic pose $\mathcal{P}_{obj}$.
    \item \textbf{Transport:} The loaded forklift must navigate through dynamic obstacles under kinematic constraints.
    \item \textbf{Precision Placement:} The cargo must be delivered to a storage slot $\mathcal{P}_{goal}$ with sub-centimeter accuracy.
\end{enumerate}
The episode terminates successfully upon satisfying the placement tolerance metric, or aborts upon collision or timeout.

\subsection{MDP Formulation}
Formally, we model this problem as a Markov Decision Process (MDP) represented by the tuple $\mathcal{M} = \langle \mathcal{S}, \mathcal{A}, \mathcal{P}, \mathcal{R}, \gamma \rangle$.

\subsubsection{State Space}
The state space $\mathcal{S}$ is high-dimensional and heterogeneous, encapsulating the multi-modal sensory data required for different operational phases. We define the composite state vector at time $t$ as $s_t = [s_{lidar, t}, s_{rgb, t}, s_{ego, t}]^T$, where:
\begin{itemize}
    \item $s_{lidar, t} \in \mathbb{R}^{N_{scan}}$ denotes the \textbf{geometric perception vector} derived from 2D laser scans, encoding the spatial occupancy required for collision-free navigation.
    \item $s_{rgb, t} \in \mathbb{R}^{H \times W \times 3}$ denotes the \textbf{semantic perception tensor} captured by the onboard RGB camera, providing dense visual cues for object detection and fine-grained alignment.
    \item $s_{ego, t} \in \mathbb{R}^{N_{prop}}$ represents the \textbf{proprioceptive state vector}, comprising the vehicle's kinematic status (linear velocity $v_t$, angular velocity $\omega_t$), actuator status (fork height $h_t$), and relative goal coordinates.
\end{itemize}

\subsubsection{Action Space}
The action space $\mathcal{A} \subset \mathbb{R}^4$ defines the continuous control inputs for the non-holonomic mobile manipulator. An action $a_t = [v_t, \omega_t, \dot{h}_t, \delta_t]^T$ corresponds to the linear velocity, steering angular velocity, fork lifting speed, and a discrete clamping trigger, respectively.

\subsubsection{Optimization Objective}
The system transitions according to the dynamics $\mathcal{P}(s_{t+1} | s_t, a_t)$, which implicitly models vehicle kinematics and contact physics. The objective is to learn a policy $\pi_\theta(a_t | s_t)$ that maximizes the expected cumulative discounted reward:
\begin{equation}
    J(\pi_\theta) = \mathbb{E}_{\tau \sim \pi_\theta} \left[ \sum_{t=0}^{T} \gamma^t R(s_t, a_t) \right]
    \label{eq:objective}
\end{equation}
where $\gamma$ is the discount factor and $R(s_t, a_t)$ is a composite reward function balancing task completion, efficiency, and safety.

\subsection{The Optimization Challenge}
Direct maximization of Eq. (\ref{eq:objective}) via a monolithic End-to-End policy is ill-posed due to \textbf{optimization interference}. The state components $s_{lidar, t}$ and $s_{rgb, t}$ serve conflicting objectives: global navigation dictates high sensitivity to $s_{lidar, t}$ for maximizing traversal speed, whereas local manipulation demands strict adherence to $s_{rgb, t}$ for minimizing alignment error. 

Compelling a single encoder to simultaneously extract features from these disparate modalities results in \textbf{gradient dominance}, where the high-magnitude gradients from collision penalties suppress the learning of subtle manipulation policies. Furthermore, the reward signal for precision placement is extremely sparse, rendering unguided exploration inefficient. These theoretical challenges motivate the \textbf{Heterogeneous Multi-Expert (HMER)} framework, which explicitly decouples the optimization landscape into specialized sub-manifolds.

\section{Heterogeneous Multi-Expert Learning System}
\label{sec:system}

To address the fundamental challenges of optimization interference and sample inefficiency inherent in long-horizon mobile manipulation, we propose the \textbf{Heterogeneous Multi-Expert Reinforcement Learning (HMER)} framework. We formally model the problem as a Semi-Markov Decision Process (SMDP), decomposing the intractable long-horizon mission into a bi-level hierarchy.

\begin{figure*}[t]
    \centering
    \includegraphics[width=0.9\textwidth]{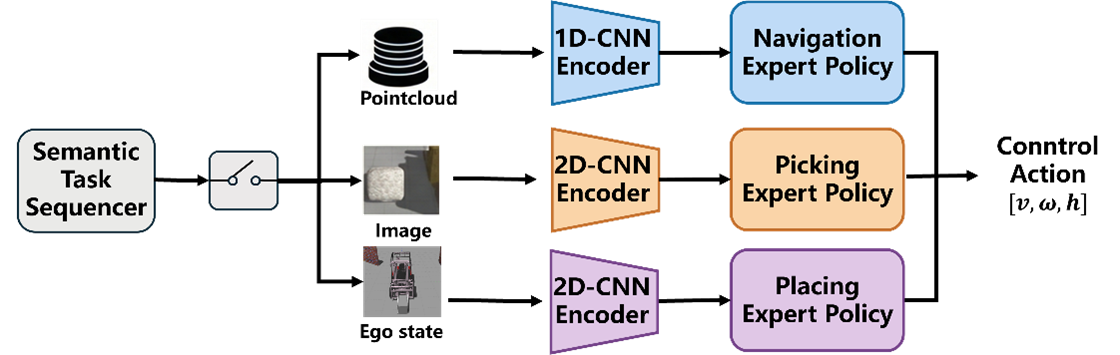}
    \caption{\textbf{The HMER Framework Architecture.} The high-level Semantic Task Planner observes discrete semantic states to orchestrate active low-level experts. Each expert is structurally decoupled, utilizing specialized encoders for distinct modalities (sparse geometric LiDAR vs. dense semantic RGB) to eliminate optimization interference.}
    \label{fig:system_arch}
\end{figure*}

The core hypothesis of HMER is that conflicting spatiotemporal requirements must be resolved through \textbf{structural decoupling} rather than joint optimization. By explicitly assigning macro-level geometric navigation and micro-level semantic manipulation to specialized modules, we isolate their respective gradient landscapes. As illustrated in the overall system architecture depicted in Fig. \ref{fig:system_arch}, our framework consists of a high-level \textbf{Semantic Task Planner} that operates on discrete events to constrain the solution space, and a set of low-level \textbf{Modality-Decoupled Expert Policies} executing high-frequency continuous control.

\subsection{Hierarchical Semantic Task Planning}
The \textbf{Semantic Task Planner}, denoted as $\pi_{H}$, serves as the cognitive core of the system, functioning as a temporal abstraction mechanism. Formally structured as a deterministic finite state automaton governing the task lifecycle, its primary role in the learning framework is to act as a \textbf{semantic manifold constrainer}.

\begin{figure}[t]
    \centering
    \includegraphics[width=0.95\columnwidth]{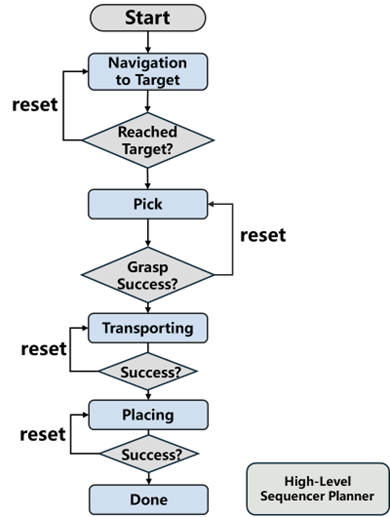}
    \caption{\textbf{Semantic Task Planner Logic.} The finite state machine governs transitions between operational phases based on semantic predicates, acting as a manifold constrainer for the learning agents.}
    \label{fig:planner_logic}
\end{figure}

Unlike end-to-end approaches that must implicitly discover task sequences from high-dimensional observations, our planner explicitly encodes the logical dependencies of the material handling cycle. As detailed in the state transition logic diagram in Fig. \ref{fig:planner_logic}, the planner observes a high-level semantic state vector $s_H$—comprising abstract features such as topological location and discrete cargo status—to activate a specific expert skill $o \in \mathcal{O}$. By conditioning expert selection on this discrete semantic state, the planner effectively masks invalid transitions, collapsing the exploration space to physically plausible sequences.

Upon activation at decision step $t_k$, the chosen expert policy $\pi_{o}$ assumes full control. The execution of skill $o$ continues until a termination condition $\beta_o(s_{t_k+k})$ is met, signaling the achievement of a sub-goal or a safety violation. This hierarchical structure bridges the gap between discrete symbolic planning and continuous sensorimotor control.

\subsection{Modality-Decoupled Expert Skills}
A critical innovation of our system is the architectural decoupling of sensory modalities at the lower level to create distinct optimization sub-spaces. Instead of a shared encoder prone to feature collapse, we design specialized neural experts with distinct encoder backbones tailored to the topology of their specific input modality.

\subsubsection{Navigation Expert (Geometric Stream)}
The navigation policy, $\pi_{nav}$, focuses exclusively on macro-level geometric perception for robust path planning in unstructured environments. Its state space input, $s_{nav}$, combines sparse 360-degree 2D LiDAR point clouds with relative polar coordinates. We design a specialized 1D Convolutional Neural Network (1D-CNN) encoder to extract spatial occupancy features from the sparse scan data. This architecture intrinsically filters high-frequency visual noise irrelevant to geometric collision avoidance. The policy outputs velocity commands optimized via a dense reward function $R_{nav}$ incentivizing efficiency and smoothness.

\subsubsection{Picking Expert (Semantic Stream)}
The picking policy, $\pi_{pick}$, governs fine-grained interaction, requiring high-resolution semantic understanding for 6-DoF alignment. Unlike navigation, its input $s_{pick}$ is derived primarily from dense RGB camera data. We utilize a deep 2D-CNN ResNet backbone to extract complex texture and edge features necessary for precise object pose estimation. The policy outputs synchronized chassis and lift control actions. To ensure robustness against approach errors, the reward $R_{pick}$ is formulated as a pose minimization problem, utilizing quaternion arithmetic to enforce precise angular alignment.

\subsubsection{Placing Expert (Precision Stream)}
The placing policy, $\pi_{place}$, addresses the ``precision bottleneck'' of industrial stacking, where sub-centimeter accuracy is mandatory. Standard distance-based reward functions suffer from vanishing gradients as the error approaches zero, halting learning near the target. To overcome this, we introduce a novel \textbf{Reciprocal Reward Formulation}:
\begin{equation}
    R_{place}(s_t) = \frac{\lambda_{prec}}{d_{target}(s_t) + \epsilon_{stab}} + R_{success}
    \label{eq:reciprocal_reward}
\end{equation}
where $d_{target}$ quantifies the combined Euclidean and angular deviation. This formulation generates increasingly sharp gradients as $d_{target} \to 0$, providing a strong, continuous optimization signal that incentivizes the agent to relentlessly refine its position to the sub-centimeter level.

\subsection{Hybrid Imitation-Reinforcement Training Strategy}
\begin{figure}[t]
    \centering
    \includegraphics[width=0.95\columnwidth]{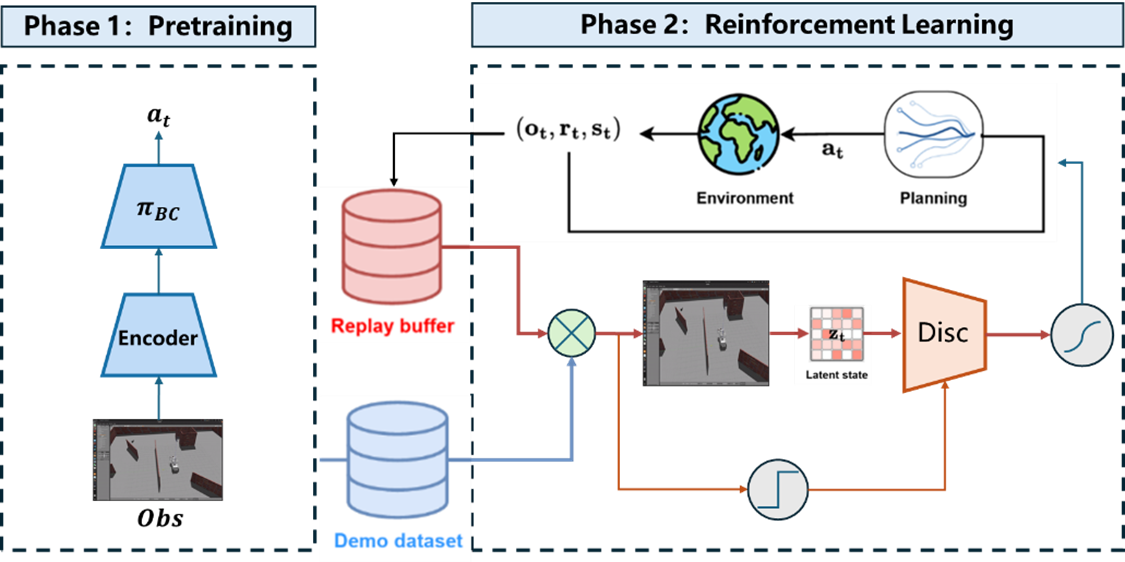}
    \caption{\textbf{Hybrid Imitation-Reinforcement Training Strategy.} Phase 1 utilizes Behavioral Cloning for stable manifold initialization. Phase 2 employs PPO for end-to-end residual refinement, enabling the experts to learn complex contact dynamics.}
    \label{fig:training_pipeline}
\end{figure}

Training this multi-expert system from scratch is non-trivial due to the extreme sparsity of the global task reward and high-dimensional state space. We introduce a two-stage \textbf{Hybrid Imitation-Reinforcement Strategy}, depicted in the training pipeline in Fig. \ref{fig:training_pipeline}, addressing the exploration-exploitation paradox in hierarchical learning.

\subsubsection{Phase 1: Manifold Initialization via Behavioral Cloning}
We first define a safe operating manifold by cloning the behavior of rule-based heuristic experts. We collect a dataset of successful demonstrations and optimize each expert policy to minimize the Negative Log-Likelihood of the expert actions. This supervised learning phase rapidly initializes the neural weights, ensuring agents possess baseline kinematic understanding and reducing early-stage catastrophic failures.

\subsubsection{Phase 2: Contact-Rich Refinement via PPO}
Heuristic demonstrations are inherently sub-optimal and fail to model complex contact dynamics essential for high-precision insertion. To surpass this baseline, we transition to Reinforcement Learning for \textbf{residual refinement}. In this phase, the pre-trained agents explore the state space around the cloned trajectories, discovering non-intuitive micro-adjustments to handle complex physical interactions. We maximize the expected cumulative reward using Proximal Policy Optimization (PPO). This phase is critical for bridging the gap between idealized heuristic planning and the physical realities of high-precision industrial manipulation.
\section{Experimental Setup and Implementation Details}
\label{sec:setup}

To ensure the reproducibility of our results and provide a rigorous benchmark for the proposed HMER framework, we detail the high-fidelity simulation environment, the specialized neural architectures designed for heterogeneous modalities, and the practical implementation protocols of the hybrid training strategy. A summary of the key implementation parameters is provided in Table \ref{tab:TRAINING}.

\begin{table}[htbp]
\centering
\setlength{\tabcolsep}{3pt} 
\caption{HYPERPARAMETERS AND SYSTEM CONFIGURATION}
\label{tab:TRAINING}
\begin{tabular}{|p{2cm}|p{3.5cm}|p{2.8cm}|}
\hline
\textbf{Category} & \textbf{Parameter} & \textbf{Value} \\
\hline
Simulation & Physics Engine & Open Dynamics Engine (ODE)\\
 & Physics Time Step ($\Delta t_{sim}$) & $0.001s$\\
 & Control Frequency & $10Hz$\\
 & Parallel Environments ($N_{env}$) & 32\\
\hline
Network Architecture & Actor/Critic Hidden Layers & [256, 256]\\
 & Activation Function & Tanh\\
 & Optimizer & Adam\\
\hline
PPO (RL Phase) & Discount Factor ($\gamma$) & 0.99\\
 & GAE Parameter ($\lambda$) & 0.95\\
 & Clip Range ($\varepsilon$) & 0.2\\
 & Entropy Coefficient (c2) & 0.01\\
 & Value Loss Coefficient (c1) & 0.5\\
 & Batch Size & 2048\\
 & Learning Rate ($\alpha_{rl}$) & $1\times10^{-4}$ (Linear Decay)\\
\hline
BC (IL Phase) & Learning Rate ($\alpha_{bc}$) & $1\times10^{-3}$\\
 & BC Epochs & 50\\
 & Expert Dataset Size & 10,000 Trajectories\\
\hline
General & Random Seed & 42\\
\hline
\end{tabular}
\end{table}

\subsection{High-Fidelity Simulation Environment}
All experimental validations are conducted within the \textbf{Gazebo 11} simulator, integrated with ROS Noetic. To bridge the ``reality gap'' and accurately model the crucial non-linear contact dynamics between the forklift tines and deformable cargo (cotton bales), we utilize the \textbf{Open Dynamics Engine (ODE)} as the physics solver with a physics time step of $\Delta t_{sim} = 0.001s$.

We implement a highly parallelized training architecture to accelerate sample collection. The environment is containerized using Docker, allowing us to instantiate $N_{env} = 32$ synchronous simulation instances on a high-performance workstation equipped with dual NVIDIA RTX 4090 GPUs and an Intel Core i9-14900K CPU. This distributed setup enables the collection of approximately $2.5 \times 10^5$ interaction steps per hour. Furthermore, to enhance the robustness of the learned policies against environmental uncertainties, we employ \textbf{Domain Randomization}: at the start of each episode, physical parameters (cargo mass, ground friction coefficients) and scene layouts (obstacle distribution) are perturbed according to predefined ranges.

\subsection{Specialized Neural Architectures}
A key feature of HMER is the use of specialized neural architectures to practically realize the \textbf{structural decoupling} introduced in Section \ref{sec:system}. The policy $\pi_\theta$ and value function $V_\phi$ for each expert are parameterized by deep neural networks tailored to the specific topology of their sensory inputs, effectively isolating gradient updates.

\subsubsection{Navigation Expert (Geometric Stream)}
The input $s_{nav}$ comprises 360-dimensional sparse LiDAR range data and a low-dimensional proprioceptive goal vector. To extract geometric features crucial for obstacle avoidance while ignoring visual noise, we process the 1D LiDAR scan using a \textbf{1D Convolutional Neural Network (1D-CNN)} consisting of three layers with kernel sizes of $[8, 4, 3]$ and strides of $[4, 2, 1]$, followed by Tanh activations. The extracted geometric embedding is concatenated with the goal vector before being fed into MLP decision heads consisting of two hidden layers containing 256 units each.

\subsubsection{Picking Expert (Semantic Stream)}
Relying on visual servoing for alignment, this input $s_{pick}$ is dominated by a $84 \times 84 \times 3$ RGB image. We employ a visual encoder based on the standard \textbf{NatureCNN} architecture, comprising three 2D convolutional layers. This structure is designed to capture rich semantic cues such as object edges and textures. The resulting flattened spatial latent features are fused with the proprioceptive fork height state.

\subsubsection{Placing Expert (Precision Stream)}
Demanding centimeter-level accuracy, the input $s_{place}$ is a low-dimensional vector representing the relative 6-DoF pose error derived from the perception system. Unlike visual agents, this compact state vector does not require convolutional processing and is fed directly into a \textbf{Multi-Layer Perceptron (MLP)} encoder to project the raw state into a dense latent embedding optimized for precise coordinate corrections.

For all experts, the specialized embeddings are fed into separate MLP networks for the Actor and Critic heads. The Actor outputs the mean $\mu$ of a diagonal Gaussian distribution, while the log-standard deviation $\sigma$ is learned as a state-independent parameter.

\subsection{Implementation of the Hybrid Training Protocol}
To ensure stable convergence in this multi-stage curriculum, we implement specific protocols for phase transition and signal normalization.

\subsubsection{Phase 1: Kinematic Manifold Initialization (BC)}
We utilize the expert dataset solely to initialize the policy network weights into a \textbf{Safe Kinematic Manifold}. The primary goal here is not perfect imitation, but to endow the agent with basic functional competency to prevent immediate collisions during early RL exploration. We employ a relatively high learning rate ($\alpha_{bc} = 1 \times 10^{-3}$) to rapidly fit the expert's geometric paths. This phase is terminated early (after 50 epochs) to prevent the policy from overfitting to the expert's inherent sub-optimality.

\subsubsection{Phase 2: Residual Refinement via PPO}
Upon transitioning to PPO for residual learning, we introduce two critical mechanisms to manage the competing optimization processes of different experts:
\begin{itemize}
    \item \textbf{Running Reward Normalization:} Since navigation tasks (measured in meters) and manipulation tasks (measured in radians or pixels) operate on vastly different scales, their raw reward signals are heterogeneous. We implement running mean-variance normalization for the rewards of each expert independently. This ensures that the advantage estimates across different skills share a similar magnitude, preventing the optimizer from biasing towards the task with larger raw reward magnitudes.
    \item \textbf{Learning Rate Annealing:} To perform fine-grained residual learning without destroying the structural knowledge acquired during the BC phase, we anneal the learning rate linearly from an initial $\alpha_{rl} = 1 \times 10^{-4}$ to zero over the course of training.
\end{itemize}

\section{Experimental Evaluation}
\label{sec:evaluation}

In this section, we present a rigorous quantitative evaluation of the proposed HMER framework. Utilizing the high-fidelity Gazebo simulation environment detailed in Section \ref{sec:setup}, we aim to answer three critical research questions:

\begin{itemize}
    \item \textbf{Q1 (Training Efficiency):} Does the hybrid imitation-reinforcement strategy effectively resolve the cold-start exploration problem and accelerate convergence compared to learning from scratch?
    \item \textbf{Q2 (Overall Performance):} Can the structurally decoupled architecture achieve superior task success rates and operational efficiency compared to monolithic or sequential baselines?
    \item \textbf{Q3 (Precision \& Robustness):} Does the system meet strict industrial standards for placement accuracy (sub-2.0 cm) and maintain robustness under environmental perturbations?
\end{itemize}

\subsection{Experimental Baselines}
We evaluate the framework in a $20m \times 20m$ simulated warehouse populated with static storage racks and dynamic obstacles. The objective is to complete a full Departure-Pick-Transport-Place cycle. To comprehensively validate the specific contributions of the HMER framework—namely the Semantic Task Planner, Modality Decoupling, and the Hybrid Training strategy—we compare it against five distinct baselines:

\begin{enumerate}
    \item \textbf{Flat-BC (Monolithic Imitation):} An end-to-end policy trained solely via Behavioral Cloning. It maps stacked raw sensor inputs (LiDAR + RGB) directly to motor actions using a single large network. This baseline highlights the impact of modality interference and the covariate shift inherent in non-hierarchical imitation.
    \item \textbf{Rule-Based (Sequential Heuristics):} A traditional modular approach executing classical algorithms in a fixed sequence: A* planning for navigation and heuristic visual servoing for manipulation. This baseline acts as the "Expert" for generating demonstration data and serves as a benchmark for adaptability and execution speed.
    \item \textbf{HBC (Hierarchical Imitation):} Incorporates the same hierarchical architecture as HMER (Semantic Planner + Experts) but is trained exclusively via Behavioral Cloning (Phase 1 only), without RL fine-tuning. This evaluates the precision bottleneck of pure supervised learning.
    \item \textbf{HRL (Hierarchical RL from Scratch):} The complete HMER architecture trained solely using PPO (Phase 2 only), without the warm-start initialization from demonstrations. This tests the necessity of the hybrid strategy in overcoming exploration sparsity.
    \item \textbf{Seq-Hybrid (w/o Semantic Planner):} A variation where the Heterogeneous Experts are trained via the full hybrid strategy but executed in a fixed, open-loop sequence without the high-level Semantic Task Planner. This validates the necessity of the semantic layer for state-aware transitions and error recovery.
    \item \textbf{HMER (Ours):} The proposed full framework featuring the Semantic Task Planner, Modality-Decoupled Experts, and the unified Hybrid Imitation-Reinforcement training strategy.
\end{enumerate}

\begin{figure}[t]
    \centering
    \includegraphics[width=0.95\columnwidth]{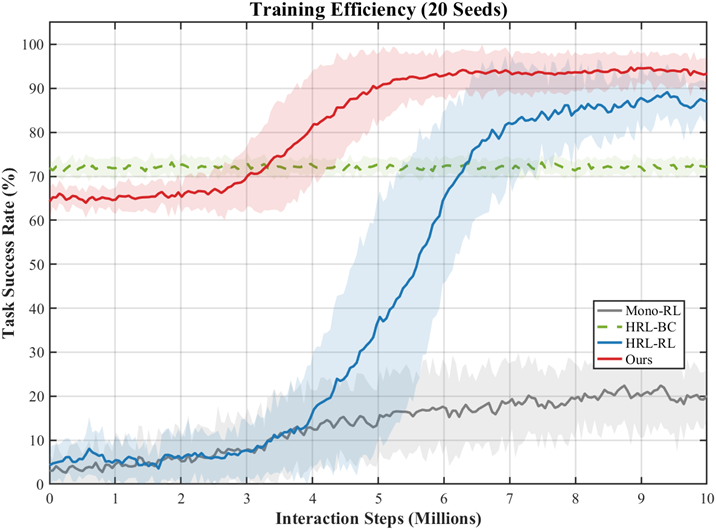}
    \caption{\textbf{Training Dynamics Analysis.} Task success rate vs. environment interaction steps. HMER (Ours) achieves faster convergence and higher asymptotic performance compared to HRL from scratch, while monolithic baselines fail to learn.}
    \label{fig:training_curves}
\end{figure}

\subsection{Training Dynamics Analysis}
We first analyze the sample efficiency of the proposed training pipeline, as illustrated in Fig. \ref{fig:training_curves} showing task success rates over 10 million interaction steps.

The results clearly demonstrate the challenge of long-horizon mobile manipulation. The Flat-BC baseline fails to generalize, stagnating below 15\% success, confirming that simple imitation cannot handle compounding errors in complex tasks. The HRL (From Scratch) agent demonstrates extremely slow convergence, requiring approximately 8.0 million steps to reach an 85\% success rate, highlighting the difficulty of exploration in sparse-reward settings.

In contrast, HMER (Ours) starts at a robust baseline of approximately 65\% (inherited from the stable kinematic manifold of the BC phase) and rapidly converges to its peak performance of 94.2\% within just 4.5 million steps. This represents a 43.7\% reduction in training samples to reach asymptotic performance compared to HRL, validating the critical role of the warm-start strategy in resolving the cold-start problem.

\subsection{Quantitative Performance Comparison}
We conducted 500 separate evaluation episodes with randomized start and target poses for each method. The comprehensive quantitative results are summarized in Table \ref{tab:main_results}.

\begin{table*}[t]
\centering
\caption{OVERALL PERFORMANCE COMPARISON (500 Evaluation Trials)}
\label{tab:main_results}
\begin{tabular}{l|c|c|c|c}
\hline
\textbf{Method} & \textbf{Task Success Rate} ($\uparrow$) & \textbf{Avg. Cycle Time} ($\downarrow$) & \textbf{Mean Placement Error} ($\downarrow$) & \textbf{Collision Rate} ($\downarrow$) \\
\hline
Flat-BC & 12.4\% & N/A & N/A & 82.5\% \\
Rule-Based (Expert) & 84.2\% & 55.6 s & 4.1 cm & 5.8\% \\
\hline
HBC (Hierarchical BC) & 76.5\% & 48.2 s & 3.9 cm & 10.2\% \\
HRL (RL from Scratch) & 88.1\% & 45.0 s & 1.8 cm & \textbf{1.5\%} \\
Seq-Hybrid (w/o Planner) & 68.5\% & \textbf{41.8 s} & \textbf{1.5 cm} & 28.4\% \\
\hline
\textbf{HMER (Ours)} & \textbf{94.2\%} & 42.5 s & \textbf{1.5 cm} & 2.1\% \\
\hline
\end{tabular}
\vspace{1ex}
\\
\raggedright \footnotesize{\textit{Note: Arrows indicate whether higher ($\uparrow$) or lower ($\downarrow$) values are better. N/A indicates the method rarely completed the task. Best results are highlighted in bold.}}
\end{table*}

\subsubsection{Task Success and Efficiency}
As shown in Table \ref{tab:main_results}, HMER achieves the highest task success rate of 94.2\%, significantly outperforming the Rule-Based expert (84.2\%) and the open-loop Seq-Hybrid baseline (68.5\%). The poor performance of Seq-Hybrid highlights the necessity of the Semantic Task Planner; without closed-loop state monitoring, minor failures in the pick phase propagate irreversibly, leading to high collision rates during transport.

Regarding operational efficiency, HMER achieves an average cycle time of 42.5s. This constitutes a 23.6\% reduction compared to the Rule-Based expert (55.6s). This efficiency gain stems from the concurrent optimization of all phases via RL: the Navigation Expert learns to cut corners and optimize path curvature better than A*, while the Manipulation Experts develop smoother, non-stop interaction policies compared to the stop-and-wait logic of visual servoing.

\subsubsection{Placement Precision and Robustness}
Terminal accuracy is the most critical metric for industrial viability. The methods relying solely on imitation (Rule-Based and HBC) plateau at a mean placement error of approximately 4.0 cm, limited by the quality of the heuristic demonstrations and sensor noise.

\begin{figure}[t]
    \centering
    \includegraphics[width=0.95\columnwidth]{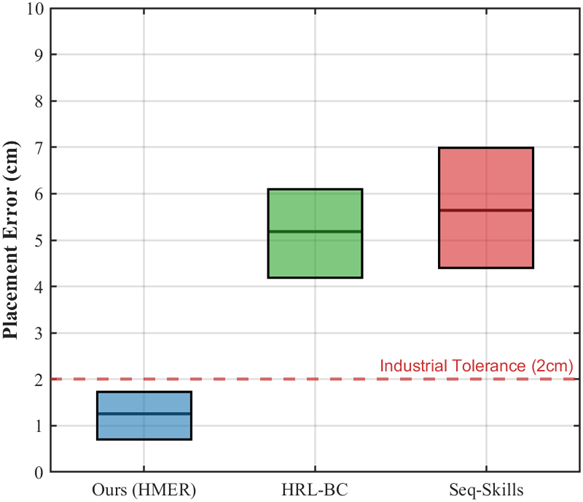}
    \caption{\textbf{Placement Accuracy Distribution.} Cumulative distribution function (CDF) of final placement error. HMER significantly outperforms imitation-based methods, achieving sub-2cm precision in the majority of trials.}
    \label{fig:error_dist}
\end{figure}

By leveraging Residual Learning with the proposed reciprocal reward, both HRL and HMER reduce this error significantly to 1.5 cm, meeting strict industrial tolerances. The error distribution analysis (referenced in Fig. \ref{fig:error_dist}) reveals that 88\% of successful HMER episodes result in an error of less than 2.0 cm, compared to only 25\% for the Rule-Based baseline. This confirms that the contact-rich refinement phase enables the agent to learn superior micro-adjustment strategies that exceed the capabilities of the supervisor.

\subsection{Ablation Analysis}
To systematically isolate the contributions of framework components, we analyze the performance drops in the ablated variants presented in Table \ref{tab:main_results}.

\textbf{Impact of Modality Decoupling:} The catastrophic failure of the Flat-BC baseline (12.4\% success) compared to the hierarchical HBC baseline (76.5\% success) provides empirical evidence for the necessity of structural decoupling. Forcing a single network to process conflicting geometric and semantic gradients leads to optimization interference, preventing the learning of even basic kinematic behaviors.

\textbf{Impact of Hybrid Training:} Comparing HBC, HRL, and HMER reveals the distinct roles of the two training phases. HBC shows that BC alone provides stable but sub-optimal capability (high precision error). HRL shows that RL alone achieves high precision but suffers from sample inefficiency and high early-stage training variance. HMER combines the best of both, utilizing BC for robust initialization and RL for precision capability, resulting in the highest overall success rate.

\textbf{Impact of Semantic Planning:} The comparison between Seq-Hybrid and HMER isolates the role of the high-level planner. While Seq-Hybrid achieves excellent precision and speed when it succeeds, its low success rate (68.5\%) and high collision rate (28.4\%) indicate brittleness. The Semantic Task Planner in HMER acts as a crucial safety layer, enabling error recovery (e.g., retrying a failed grasp instead of proceeding to transport), which is essential for robustness in unstructured environments.

\section{Conclusion}
\label{sec:conclusion}
Future work will focus on two main directions. We aim to transfer the learned policies to a real-world forklift platform using Sim-to-Real techniques such as Domain Randomization and Domain Adaptation. Additionally, we plan to extend the high-level policy to handle multi-agent coordination, enabling a fleet of autonomous forklifts to collaborate efficiently in dynamic, shared workspaces.
In this work, we addressed the fundamental challenge of autonomous mobile manipulation in unstructured warehouses, where the conflicting requirements of large-scale navigation and high-precision stacking create a significant barrier for monolithic learning approaches. We proposed a Collaborative Multi-Expert Reinforcement Learning (CMRL) framework tailored for autonomous forklifts. By decoupling the sensory processing streams, we successfully resolved the "Modality Interference" problem, enabling a geometry-aware Navigation Expert and semantics-aware Manipulation Experts to cooperate seamlessly under a semantic task sequencer. Furthermore, our "Imitate-then-Reinforce" hybrid training strategy proved effective in solving the exploration-precision trade-off. We demonstrated that while Imitation Learning provides a safe kinematic initialization, the subsequent Residual Learning phase is indispensable for mastering the contact-rich dynamics required for industrial-grade precision.
Extensive evaluations in a high-fidelity Gazebo simulation validated the efficacy of our approach. The CMRL system achieved a task success rate of 94.2 \%, significantly outperforming end-to-end baselines (12.4\%) which failed to converge due to gradient conflicts. Most notably, our method broke the precision ceiling of heuristic experts, reducing the placement error to 1.5 cm and meeting strict industrial tolerances. The system also demonstrated superior operational efficiency, reducing the cycle time by 21.4\% compared to sequential execution baselines, and exhibited robust performance even under severe environmental perturbations.
Despite these promising results, we acknowledge several limitations. First, our validation is currently restricted to simulation. While we modeled physics fidelity carefully, real-world factors such as sensor latency and uneven floor friction may introduce discrepancies. Second, the current high-level sequencer operates on a discrete logic basis; it does not yet possess the capability to learn continuous, creative recovery strategies outside its pre-defined semantic transitions.
Future work will focus on two primary directions. To bridge the Sim-to-Real gap, we plan to implement domain adaptation techniques to transfer the visual encoders to physical hardware. Additionally, we aim to extend the framework to Multi-Agent Reinforcement Learning, coordinating a fleet of autonomous forklifts to optimize global warehouse throughput while managing traffic congestion in shared workspaces.
\bibliographystyle{IEEEtran}
\bibliography{reference.bib}

\end{document}